\pdfoutput=1

\documentclass[11pt]{article}

\usepackage[preprint]{acl}

\usepackage{times}
\usepackage{latexsym}

\usepackage[T1]{fontenc}

\usepackage[utf8]{inputenc}

\usepackage{microtype}

\usepackage{inconsolata}

\usepackage{graphicx}
\usepackage{multirow}
\usepackage{amsmath}
\usepackage{enumitem}

\title{Large Language Model-Enhanced Symbolic Reasoning for Knowledge Base Completion}

\author{Qiyuan He\textsuperscript{1}
        \hspace{3mm}
        Jianfei Yu\textsuperscript{2}
        \hspace{3mm}
        Wenya Wang\textsuperscript{1}
        \\
        \textsuperscript{1}College of Computing and Data Science, Nanyang Technological University\\
        \textsuperscript{2}School of Computer Science and Engineering, Nanjing University of Science and Technology
        \\
        \texttt{\{qiyuan001,wenya\}@ntu.edu.sg}
        \hspace{3mm}
        \texttt{jfyu@njust.edu.cn}
}

\begin{document}
\maketitle

\begin{abstract}
Integrating large language models (LLMs) with rule-based reasoning offers a powerful solution for improving the flexibility and reliability of Knowledge Base Completion (KBC). 
Traditional rule-based KBC methods offer verifiable reasoning yet lack flexibility, while LLMs provide strong semantic understanding yet suffer from hallucinations. 
With the aim of combining LLMs' understanding capability with the logical and rigor of rule-based approaches, we propose a novel framework consisting of a Subgraph Extractor, an LLM Proposer, and a Rule Reasoner. The Subgraph Extractor first samples subgraphs from the KB. Then, the LLM uses these subgraphs to propose diverse and meaningful rules that are helpful for inferring missing facts. To effectively avoid hallucination in LLMs' generations, these proposed rules are further refined by a Rule Reasoner to pinpoint the most significant rules in the KB for Knowledge Base Completion.
Our approach offers several key benefits: the utilization of LLMs to enhance the richness and diversity of the proposed rules and the integration with rule-based reasoning to improve reliability. Our method also demonstrates strong performance across diverse KB datasets, highlighting the robustness and generalizability of the proposed framework.\footnote{This work will be submitted to the IEEE for possible publication. Copyright may be transferred without notice, after which this version may no longer be accessible.}
\end{abstract}

\section{Introduction}
Knowledge bases (KBs) are repositories of structured information that serve foundational roles in a wide range of machine learning applications, such as question-answering, recommendation systems, and semantic search \cite{wang2017knowledge}. Despite their compact and large volume of information storage, KBs are often incomplete, leading to significant gaps in knowledge representation. To tackle this challenge, the task of Knowledge Base Completion (KBC) has attracted considerable attention in the research community, aiming to automatically infer missing entities within KBs \cite{socher2013reasoning}.

\begin{figure}[!tp]
\centering
\includegraphics[width=0.95\linewidth,trim=0cm 0.8cm 0cm 0cm]{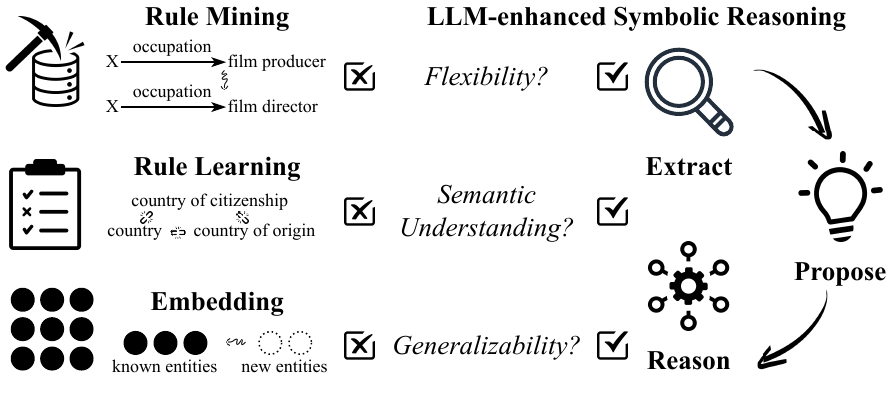} 
\caption{\textbf{LeSR}: \textbf{L}LM-\textbf{e}nhanced \textbf{S}ymbolic \textbf{R}easoning for KB Completion, aiming for flexibility, semantic understanding and generalizability.
}
\label{figure: motivation teaser}
\end{figure}

Existing KBC methods are mostly embedding-based and logic rule-based. 
Earlier research focuses on embedding-based methods that learn to encode the semantics of entities and relations as vectors, enabling efficient inference through vector operations \cite{Bordes2013TranslatingEF, Yang2014EmbeddingEA, Trouillon2016ComplexEF, balazevic-etal-2019-tucker}. These methods have gained popularity due to their effectiveness in capturing latent patterns within large-scale KBs.
On the other hand, logic rule-based methods \cite{Khot2011LearningML, Rocktschel2017EndtoendDP, Yang2017DifferentiableLO, Qu2020RNNLogicLL} offer an interpretable approach to KBC, contrasting with the often opaque nature of embedding-based methods \cite{Bordes2013TranslatingEF, Yang2014EmbeddingEA, Trouillon2016ComplexEF, balazevic-etal-2019-tucker}. 
They leverage logic rules such as $parent(A, B) \wedge parent(B, C) \Rightarrow grandparent(A, C)$, which reads as \textit{if A is B's parent and B is C's parent, then A is C's grandparent}, to infer missing facts based on existing KB.

Nevertheless, most existing KBC methods still suffer from several limitations.
Embedding-based methods, while efficient, often lack interpretability and fail to handle new entities unseen during training. Although rule-based methods excel in providing transparent and verifiable reasoning, either through pattern mining \cite{Galrraga2013AMIEAR,Wang2015RDF2RulesLR, Meilicke2019AnytimeBR} or neural modeling \cite{Rocktschel2017EndtoendDP, xiong-etal-2017-deeppath, Minervini2020LearningRS}, they struggle to
identify quality and diverse rules. 
The lack of flexibility and diversity limits the effectiveness of these rule-based approaches due to the increasing complexity and scale of modern KBs \cite{Zhou2023ReliableKG}.

To tackle the aforementioned limiations, we consider to leverage large language models in the rule mining process and integrate the power of LLMs into the symbolic reasoning process of KB completion. With vast linguistic knolwedge captured from large-scale pre-training, LLMs have been used for KBC by framing it as a sequence generation problem \cite{Yao2023ExploringLL} where LLMs are used to directly infer missing entities or relationships \cite{Yao2023ExploringLL} given a query such as ``\textit{What is the capital of France?}''.
However, treating KBC as a generation task with LLM backbones has raised concerns regarding transparency and accuracy. Namely, these approaches heavily rely on the inherent abilities of LLMs, lacking clarity in the internal reasoning processes. Moreover, LLMs are prone to hallucinations and errors and tend to perform poorly without extensive fine-tuning, especially on domain-specific knowledge \cite{Veseli2023EvaluatingLM, Zhang2023MakingLL, He2024LinkGPTTL}.

We propose a novel framework, \textbf{LeSR} (\textbf{L}LM-\textbf{e}nhanced \textbf{S}ymbolic \textbf{R}easoning), which synergizes the comprehensive understanding capabilities of LLMs and the rigorousness of rule-based systems. LeSR consists of a Subgraph Extractor, an LLM Proposer and a Rule Reasoner, designed to enhance the relevancy and diversity of logic rules and maximize the effectiveness and reliability of the knowledge inference process.
For each relation in the KB, the Subgraph Extractor is responsible for identifying meaningful subgraphs surrounding the relation, which will be further fed into the LLM Proposer to generate diverse and relevant logic rules. The comprehensive power of LLMs contributes to identifying a wide range of entity-agnostic logic rules, uncovering common patterns behind the extracted subgraphs, but meanwhile, it also brings in unexpected noise, which is harmful to knowledge completion. We thus introduce a Rule Reasoner to refine LLM proposals by learning to score each rule, improving reliability and reducing the hallucination inherent in the LLM Proposer.

The contributions of our work are threefold: (1) We introduce a novel paradigm leveraging LLMs to propose logic rules that are relation-specific and sensitive to subgraph structures. The proposed rules demonstrate sufficient diversity and coverage. (2) We propose a novel framework that effectively integrates LLMs with logic rule reasoning. Our framework combines the power of language understanding and rigorous reasoning, leading to a transparent and more reliable inference process, effectively mitigating the errors produced by LLMs alone. (3) We conduct extensive experiments over five knowledge base benchmarks, covering diverse domains and complexities. Our method achieves comparable results across all datasets and produces adequate interpretable rules, demonstrating its effectiveness as a robust and generalizable solution for KBC.

\begin{figure*}[!htp]
\centering
\includegraphics[width=0.95\textwidth,trim=0cm 0cm 0cm 0cm]{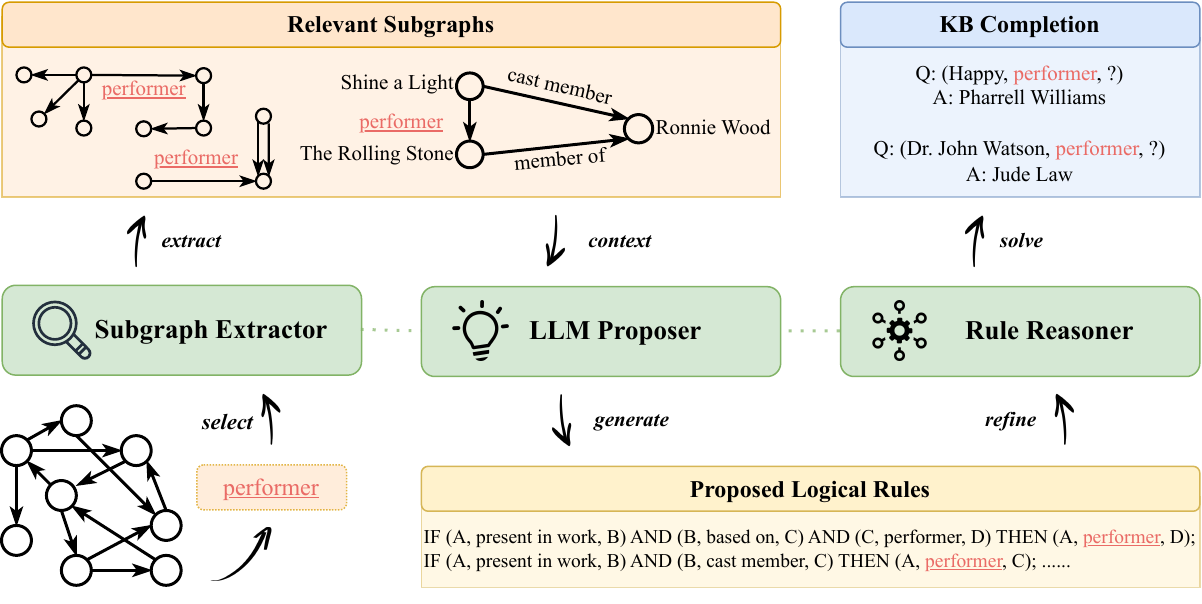} 
\caption{An overview of \textbf{LeSR}: LLM-enhanced Symbolic Reasoning. The Subgraph Extractor samples relevant subgraphs from the KB, then the LLM uses these subgraphs to propose logical rules, which will be further refined by the Rule Reasoner, learning the significance of the proposed rules and performing KB completion.
}
\label{figure: overview of LeSR Model}
\end{figure*}

\section{Related Work}

Knowledge Base Completion (KBC), sometimes known as link prediction, is the task of filling in missing information in knowledge bases (KBs) based on existing data. Existing KBC methods can be broadly and loosely categorized as embedding-based or rule-based strategies.

\paragraph{Embedding-based KBC} Most embedding-based methods represent entities and relations by vectors with their semantics preserved in the embedding space \cite{Sun2018RotatEKG, Zhang2022KnowledgeGR, Bordes2013TranslatingEF, Trouillon2016ComplexEF, Dettmers2017Convolutional2K, balazevic-etal-2019-tucker}. They are black-box in nature and thus lack interpretability. In addition, they heavily rely on data of good quality to excel \cite{Nickel2011ATM}. Furthermore, these embedding-based methods implicitly follow the closed world assumption \cite{Qu2020RNNLogicLL, Paulheim2016KnowledgeGR} that all facts not present in the knowledge base dataset are false. However, under real-world scenarios, knowledge bases tend to be inherently incomplete and follow the open-world assumption, i.e. the absence of a fact implies uncertainty instead of falsehood. Because closed-world assumption does not account for the possibility of the unknown, KBC methods based on these methods suffer from incomplete or evolving KBs.

Some works use graph neural networks (GNN) to solve KBC. In general, GNN-based approaches learn embeddings for both entities (nodes) and relationships (edges) in the KB and use message-passing to aggregate information from neighbouring nodes and edges \cite{Schlichtkrull2017ModelingRD}. However, GNNs may quickly find and optimize for existing link existence information in the training data, potentially leading to overfitting and poor generalizability \cite{Zhang2018LinkPB}.

\paragraph{Rule-based KBC} Rule-based methods assume relationships between entities, and relations can be explicitly expressed as logical rules or patterns \cite{ Nickel2015ARO}. Early representative methods include rule mining \cite{Meilicke2019AnytimeBR}, markov logic networks \cite{Khot2011LearningML}, relational networks \cite{Natarajan2010BoostingRD}, neuro symbolic models \cite{Yang2019LearnTE, Sadeghian2019DRUMED, Qu2020RNNLogicLL} and neural theorem provers \cite{Minervini2020LearningRS, Rocktschel2017EndtoendDP}. While offering verifiable reasoning, they rely on extensive searching, mining, and application of rules across KBs that are computationally intensive \cite{Zeng2023LogicalRK}. In addition, the rules may be tailored to specific relationships or patterns in the KB, limiting their ability to generalize to new or unseen scenarios \cite{Wu2023RuleLO}.

Another paradigm adopts reinforcement learning to learn rules \cite{Das2017GoFA, lin-etal-2018-multi}. These methods model the KB reasoning process as a sequential decision-making problem and treat KBC as a Markov Decision Process, where an RL agent explores paths in the KB to validate facts. 
However, training effective path-finding agents is particularly difficult due to the sparsity of the reward signal, making them underperform compared to other alternatives \cite{Qu2020RNNLogicLL}. 
Moreover, RL agents often struggle to navigate the large search space to identify optimal paths, leading to reliability issues in KB reasoning \cite{Zhou2023ReliableKG}.

\paragraph{LLMs for KBC}
There also exist some works utilizing LLMs for Knowledge Base Completion. The general idea is to treat KBC as a language generation task and use LLMs to generate answers to the query. 
However, KBC is generally considered knowledge-intensive tasks that require a significant amount of external knowledge as a supplement, and existing works focused either on prompt design or fine-tuning LLMs for performance gain \cite{Veseli2023EvaluatingLM, Zhang2023MakingLL, Yao2023ExploringLL, He2024LinkGPTTL}.
Notably, commonsense KBs \cite{sap2019atomicatlasmachinecommonsense, Speer2016ConceptNet5A} differ from domain-specific KBs as they are inherently very sparse and incomplete. Some research uses LLMs for commonsense KB completion and construction with a focus on capturing the implicit knowledge not reflected in the commonsense KBs \cite{Hwang2020COMETATOMIC2O, bosselut-etal-2019-comet, lin-etal-2019-kagnet}.

\section{Method}

We propose a novel framework, \textbf{LeSR}, consisting of a Subgraph Extractor, an LLM Proposer, and a Rule Reasoner, as shown in Fig.~\ref{figure: overview of LeSR Model}. 
The Subgraph Extractor samples a set of relevant subgraphs $\mathcal{G}_r$ surrounding each relation $r$ in the knowledge base (KB), providing context-rich and structure-preserved inputs to the LLM. The LLM then uses these subgraphs to generate diverse and meaningful rules $\Phi_{r}$, uncovering common patterns inherent with these input subgraphs. Each rule $\phi\in \Phi_{r}$ is a logic clause in the form of IF-THEN format, consisting of knowledge triplets as predicates (e.g., $(A, r, B)$ with variable $A$ and $B$ as placeholders for arbitrary entity $h,t$ and relation $r$).
A simple example of $\phi$ can be ``IF $(A, r_1, B)$ AND $(B, r_2, C)$ THEN $(A, r, C)$'', which is read as ``\textit{If entity A has relation $r_1$ to B, and entity B has relation $r_2$ to C, then A has relation $r$ to C.}'' The generated rules are subsequently refined and scored by a Rule Reasoner during training to improve their reliability and reduce the likelihood of error and hallucination.

\subsection{Problem Definition}
Knowledge Base Completion refers to the task of inferring missing facts in a knowledge base. A knowledge base is typically represented as a set of triplets $(h,r,t)$, where $h$ (head) and $t$ (tail) are entities, and $r$ (relation) represents the relationship between them. In this sense, a KB can be considered as a directed graph with nodes being entities and edges being relations. The goal of KBC is to predict new triplets $(h,r,t)$ that are likely to be true but are not currently present in the KB, thereby completing the knowledge base.

Let $\mathcal{E}$ be the set of all entities and $\mathcal{R}$ be the set of all relations in the knowledge base. The knowledge base is then defined as a set of observed triples:
$\mathcal{G} = \{(h,r,t)\ |\ h,t \in \mathcal{E}, r\in \mathcal{R}\}$.
The KBC task is to find the most likely $t$ given an incomplete triplet $(h,r,?)$ missing the object. This task can be formulated as finding $t = \arg\max_{t'} f(h,r, t')$, where $f$ is the scoring function that models the plausibility of the triplet $(h,r, t')$.

\subsection{Subgraph Extractor} 
To maximize the effectiveness of LLMs in generating candidate rules, it is essential to integrate useful information pertinent to a specific relation into the input context of the LLMs. Given a training KB, subgraphs surrounding a specific relation $r$ serve as a natural source of structure-aware context to be fed into LLMs. Given the diverse coverage of the training KB, we expect that a set of different subgraphs for $r$ sampled from the KB is able to reflect both diverse and common patterns to inform the set of plausible logic rules leading to the target relation $r$.
Concretely, the Subgraph Extractor is designed to output a set of representative subgraphs $\mathcal{G}_r = \{g_r | g_r\subset \mathcal{G}\}$ centring each relation $r\in\mathcal{R}$. 

We consider all triplets $(\cdot, r, \cdot)$ in $\mathcal{G}$ with relation $r$ and perform random sampling to sample $m$ triplets to create a set of target triplets of $r$: $\{(h_1, r, t_1), ..., (h_m, r, t_m) \}$.
Then for each target triplet $(h_j, r, t_j)$, we perform a multi-hop traversal starting from its head and tail entities $h_j, t_j$. Each hop collects all adjacent entities with an edge directly connected to the pivot entity. 
The collection of triplets at the $(k+1)$-hop pivoting at triplet $(h_j,r,t_j)$ is derived as:
$N^{k+1}_{(h_j, r, t_j)} = \{(h,r,t)\in\mathcal{G} | h\in E^{k}_{(h_j, r, t_j)} \textrm{ or } t\in E^{k}_{(h_j, r, t_j)}\}$ where $E^{k}_{(h_j, r, t_j)}$ represents the entities appeared in $N^{k}_{(h_j, r, t_j)}$ which is the $k$-hop traversal subgraph of target triplet $(h_j, r, t_j)$.
To ensure the compactness of the sampled subgraph, we further require that the triplets visited during the last hop of traversal are directly adjacent to the target triplet. In this way, the obtained subgraph ${g_r}_j$ contains neighbouring entities with multiple closed paths from the target triplet.

\subsection{LLM Proposer} 
The LLM Proposer is responsible for generating diverse and meaningful logic rules for KBC. It is designed to leverage the extensive linguistic capabilities of LLMs to create rules that are tailored to target relations within the provided knowledge subgraph. This module improves the flexibility and diversity of the rules compared to traditional rule-based or neural methods.

More specifically, given the set of sampled subgraphs $\mathcal{G}_r$ corresponding to each relation $r\in\mathcal{R}$ as discussed in the previous section, we linearize each subgraph into a textual prompt to be fed as the LLM input. The prompt is designed so that the output is of the form ``IF X THEN Y,'' where $X$ represents the rule body and $Y$ the rule head, potentially including conditions with varying logical structures \footnote{While LLMs are capable of generating more complex logic rules involving disjunctions (OR) and negations (NOT) like ``IF (A, headquarters location, B) OR (A, capital, B) THEN (A, country, B)" and ``IF (A, shares border with, B) AND NOT (B, shares border with, C) THEN NOT (A, shares border with, C)", in this work, we focus exclusively on conjunctions (i.e., AND conditions). This decision simplifies the rule induction process while still achieving high-quality predictions.}, and the template used for generating these prompts can be found in Appx.~\ref{appendix: prompt templates}.  The LLM is prompted to generate the most relevant logical rules that can be used to induce the target relation $r$. We denote by $\Phi_r = {\phi_1, ..., \phi{n_r}}$ the final output from the LLM Proposer consisting of candidate logic rules where each rule $\phi_i$ can be logically derived as $\textrm{X} \Rightarrow \textrm{Y}$.

As LLMs may sometimes produce imprecise outputs, to obtain syntactically meaningful rules $\Phi_{r}$, we refine the generated rules from LLMs via a two-stage filtering process. 
In the first stage, we filter out rules that do not adhere to the correct format to ensure proper logical structure (e.g. incomplete statements with missing rule heads), rules with irrelevant rule heads (e.g. complete statements but for rules other than the given relation, and rules that are not entity-agnostic (e.g. rules that reference specific entities rather than generalizable concepts).
In the second stage, we further process the filtered rules by mapping all the relations in each rule to their corresponding KB relations in $\mathcal{R}$ via embedding-based semantic similarity. For example, given a generated rule $\phi$ expressed as ``IF $(A, r_1, B)$ AND $(C, r_2, B)$, THEN $(B, r_3, C)$'', we use the sentence transformer \cite{reimers-gurevych-2019-sentence} to map $r_i, i\in\{1,2,3\}$ to $r'_i\in\mathcal{R}$ of the highest similarity score for semantic flexibility.
In this way, we take advantage of the linguistic understanding of LLMs to obtain a set of flexible and diverse logical rules $\Phi_r=\{\phi_{1}, ..., \phi_{n_{r}}\}$ proposed by LLM for each relation $r\in\mathcal{R}$, where $n_r$ is the number of the total logic rules for relation $r$. 

\subsection{Rule Reasoner} 
The Rule Reasoner evaluates, refines, and scores the logic rules proposed by the LLM in the previous stage. It is critical to ensure that these proposed rules are not only plausible but also reliably grounded to the KB, reducing the likelihood of errors or hallucinations that LLMs might introduce. To effectively score the generated logic rules, we propose to ground the general rules in the given knowledge base (\emph{Rule Grounding}) and evaluate their relevancy 
in the KB (\emph{Rule Evaluation}).

\paragraph{Rule Grounding}
For each logic rule $\phi_i$ for relation $r\in\mathcal{R}$, we perform knowledge base grounding, a process that maps abstract logical rules to specific entities and relationships within the given KB, to identify corresponding pairs of $h,t\in\mathcal{E}$ so that $(h,r,t)\in\mathcal{G}$. This process produces two matrices $\mathbf{C}_i$ and $\mathbf{A}_i$, where $\mathbf{C}_i$ records the pairs of $(h,t)$ entities that satisfy the rule body of $\phi_i$, and $\mathbf{A}_i$ records the pairs of $(h,t)$ that satisfy both the rule body and rule head of $\phi_i$, and thus making $\phi_i$ successfully grounded. For example, given a logic rule $\phi=$``IF $(A, parent, B)$ AND $(B, parent, C)$ THEN $(A, grandparent, C)$'' and a few KB facts: (Anna, parent, Bob), (Bob, parent, Charlie), (Anna, grandparent, Charlie), we say that (Anna, Charlie) satisfy the entire rule $\phi$, whereas (Anna, Bob) and (Bob, Charlie) satisfy the body of $\phi$.

Mathematically, $\mathbf{C}_i = [c_i(h,t)]_{|\mathcal{E}|\times |\mathcal{E}|} $ with $c_i(h,t)$ being the number of traversing options to traverse over $\mathcal{G}$, starting from entity $h$, following the rule body of $\phi_i$, and ending at entity $t\in\mathcal{E}$. $c_i(h,t)=0$ if there is no traversing options between $h$ and $t$ following $\phi_i$. Similarly, we have $\mathbf{A}_i = [a_i(h,t)]_{|\mathcal{E}|\times |\mathcal{E}|} $ with $a_i(h,t)$ being the number of available traversal paths over $\mathcal{G}$ to traverse from entity $h$ to reach entity $t$, following $\phi_i$'s rule body and rule head.
We categorize logic rules based on their complexity and traversal structures, e.g. 0th-order inversion $(A, r_i, B) - (B, r_j, A)$ and 1st-order bidirectional $(B, r_i, A) - (B, r_j, C) - (A, r_k, C)$.
As each relation $r$ corresponds to a matrix $M_{r} = \{m_{ht}\}_{|\mathcal{E}|\times |\mathcal{E}|}$ where $m_{ht}=1$ if $(h,r,t)\in\mathcal{G}$ and $m_{ht}=0$ otherwise, we can use matrix multiplications to represent the conjunctions of atomic triplets and matrix transpose to represent an inversion.
The detailed categorization of logic rules and their corresponding formula to compute $\mathbf{C}$ and $\mathbf{A}$ can be found in Appx.~\ref{appendix: rule traversal structures}.

\paragraph{Rule Evaluation}
We can further define a scoring function $s_i$ using $\mathbf{A}_i$ and $\mathbf{C}_i$ as a measurement of the grounding quality for an arbitrary $(h,r,t)\in\mathcal{G}$:
$s_i(h,t)=\mathbf{A}_i(h,t)$ when $\mathbf{A}_i(h,t)>0$, $s_i(h,t)= -\mathbf{C}_i(h,t)$ when $ \mathbf{A}_i(h,t)=0 \wedge \mathbf{C}_i(h,t)>0$ and $s_i(h,t)=0$ otherwise.

The main idea is to award when $\phi_i$ can be used to obtain correct $t$ and penalize if such grounding leads to an incorrect entity. Here $s_i(h,t)$ measures the grounding quality of $\phi_i$ in the entity pair $(h,t)$ with regard to the target relation $r$.
From the perspective of KBC, answering query $(h,r,?)$ with logic rule $\phi_i$ is to find $t'$ that maximizes $f_{\phi_i}(h,r,t') = s_i(h,t')$. We associate each logical rule $\phi_i$ with a learnable significance score $w_i$ and optimize these $w_i$ using $L_0 = -\log \sum_{\phi_i} w_i f_{\phi_i}$.

To account for cases which are not covered by logic rules, we additionally incorporate an embedding-based model RotatE \cite{Sun2018RotatEKG} that calculates a score $f_{emb}(h,r,t')=RotatE(h,r,t')$ based on semantic information. The training object of the \textbf{LeSR} model is to learn optimal rule significance $w_i$, $w_{emb}$ and weighting factor $\alpha$ for the following loss function: 
$$L_1= -\log \alpha \sum_{\phi_i} w_i f_{\phi_i} + (1-\alpha) w_{emb} f_{emb}$$
Due to the sparsity of KBs, we focus on nonzero contributions by masking zero $s_i$ during training. We also use softmax transformation to $w_i$ and $w_{emb}$ to ensure they are positive and sum to one.

During inference, the model process query $(h,r,?)$ and computes the weighted likelihood of each $t'\in\mathcal{E}$ being the answer to this query using logical rules and embedding. Entities are ranked based on their weighted likelihood, allowing us to compute the rank of the correct answer entity.

\section{Experiments}

\subsection{Experimental Setup}

\paragraph{Datasets} We choose five KB benchmarks, namely (1) \textbf{UMLs}, a biomedical KB \cite{Kok2007StatisticalPI} with many domain-specific entities; (2) \textbf{WN18RR}, an English Lexical-focused dataset derived from WordNet \cite{Dettmers2017Convolutional2K} with very few relations; (3) \textbf{CN100K}, a commonsense KB derived from ConceptNet \cite{Speer2016ConceptNet5A} proposed by \citet{li-etal-2016-commonsense}; (4) \textbf{FB15K-237}
, a general KB derived from Freebase \cite{toutanova-chen-2015-observed}; and (5) \textbf{WD15K}, another general KB based on Freebase with additional annotation on rule quality \cite{lv-etal-2021-multi}. 
As summarized in Tab.~\ref{table: data_stat}, these datasets span multiple domains with KBs of varying graph connectivities and sparsity levels, ensuring diversity.

\begin{table}[htp]
\centering
\scriptsize  
\setlength{\tabcolsep}{1mm}
\begin{tabular}{c|ccccc}
\hline
\multirow{2}{*}{Statistic} & \multicolumn{5}{c}{Dataset}                                 \\
                      & UMLs       & WN18RR     & FB15K-237 & WD15K   & CN100       \\ \hline
\#Train                & 1,959      & 86,835     & 272,115   & 159,036 & 100,000     \\
\#Valid                & 1,306      & 3,134      & 17,535    & 8,727   & 1,200       \\
\#Test                & 3,264      & 3,034      & 20,466    & 8,761   & 1,200       \\
\hline
\#Entity              & 135        & 40,943     & 14,541    & 15,812  & 78,339      \\
\#Relation             & 46         & 11         & 237       & 179     & 34          \\ \hline
Sparsity                & 36.100\%   & 0.005\%    & 0.147\%   & 0.071\% & 0.002\%     \\
Avg degree               & 48.60      & 2.27       & 21.33     & 11.16   & 1.31 \\
\hline
\end{tabular}
\caption{Dataset statistics for KGs in use}
\label{table: data_stat}
\end{table}

\paragraph{Metrics} \label{paragraph: metrics}

We follow \citet{Galkin2022InductiveLQ} to use the standard train/test/valid partitioning of the complete KB dataset to create the training, testing, and validation KBs. 
We use the standard evaluation metrics for KBC: mean rank (MR), mean reciprocal rank (MRR) and Hit@K with K=1,3,10 reported in percentage.

\begin{table*}[htp]
\centering
\small  
\setlength{\tabcolsep}{1.5mm}

\begin{tabular}{c|c|ccccc|ccccc}
\hline
\multirow{2}{*}{Category}   & \multirow{2}{*}{Model} & \multicolumn{5}{c|}{UMLs}                  & \multicolumn{5}{c}{WN18RR}                  \\
 &
   &
  MR $\downarrow$ &
  MRR$\uparrow$ &
  H@1$\uparrow$ &
  H@3$\uparrow$ &
  H@10$\uparrow$ &
  MR $\downarrow$ &
  MRR$\uparrow$ &
  H@1$\uparrow$ &
  H@3$\uparrow$ &
  H@10$\uparrow$ \\ \hline
Non-ML                      & AnyBURL                & -   & 0.729 & 62.5          & 79.5 & 93.5  & -      & 0.498 & 45.8 & 51.2          & 57.7 \\ \hline
Emb.                        & RotatE                 & 4.1 & 0.731 & 62.4          & 80.8 & 93.3  & 3365.3 & 0.476 & 42.8 & 49.6          & 57.5 \\ \hline
\multirow{3}{*}{Rule} &
  NeuralLP &
  4.0 &
  0.735 &
  60.4 &
  \underline{84.9} &
  91.7 &
  \textbf{43.3} &
  \underline{0.509} &
  \underline{46.2} &
  49.9 &
  \underline{62.0} \\
 &
  DRUM &
  \textbf{2.2} &
  \textbf{0.800} &
  65.2 &
  \textbf{94.5} &
  \textbf{97.2} &
  \underline{43.1} &
  \textbf{0.515} &
  \textbf{47.1} &
  50.2 &
  \textbf{62.1} \\
                            & RNNLogic               & 8.0 & 0.655 & 53.7          & 72.5 & 86.9  & 5983.3 & 0.426 & 39.9 & 43.4          & 47.9 \\ \hline
GNN                         & RGCN                   & -   & 0.434 & 26.5          & 53.0 & 76.0  & -      & 0.359 & 31.5 & 38.5          & 43.1 \\ \hline
\multirow{2}{*}{GPT-3.5}    & Inference              & -   & -     & 1.2           & 3.2  & 11.6  & -      & -     & 4.5  & 19.9          & 28.2 \\
                            & LeSR                   & 4.1 & 0.764 & \textbf{68.2} & 81.5 & 91.8  & 1989.0 & 0.497 & 44.0 & \textbf{52.3} & 61.0 \\ \hline
\multirow{2}{*}{GPT-4}      & Inference              & -   & -     & 1.04          & 3.74 & 19.09 & -      & -     & 0.14 & 0.28          & 0.35 \\
 &
  LeSR &
  \underline{3.8} &
  \underline{0.769} &
  \underline{67.7} &
  83.1 &
  \underline{94.2} &
  1976.8 &
  0.326 &
  22.6 &
  36.4 &
  55.4 \\ \hline
\multirow{2}{*}{Gemini-1.5} & Inference              & -   & -     & 0.9           & 4.1  & 18.1  & -      & -     & 10.1 & 26.8          & 34.3 \\
                            & LeSR                   & 6.3 & 0.723 & 65.8          & 75.8 & 83.2  & 1987.2 & 0.489 & 42.9 & \underline{51.7} & 60.9 \\ \hline
\end{tabular}
\caption{KBC performance on UMLs and WN18RR. 
\label{table: results_12}
}
\end{table*}

\begin{table*}[htp]
\centering
\small  
\setlength{\tabcolsep}{1.5mm}

\begin{tabular}{c|c|ccccc|ccccc}
\hline
\multirow{2}{*}{Category} &
  \multirow{2}{*}{Model} &
  \multicolumn{5}{c|}{FB15K-237} &
  \multicolumn{5}{c}{WD15K} \\
 &
   &
  MR $\downarrow$ &
  MRR$\uparrow$ &
  H@1$\uparrow$ &
  H@3$\uparrow$ &
  H@10$\uparrow$ &
  MR $\downarrow$ &
  MRR$\uparrow$ &
  H@1$\uparrow$ &
  H@3$\uparrow$ &
  H@10$\uparrow$ \\ \hline
Non-ML &
  AnyBURL &
  - &
  0.319 &
  23.8 &
  34.9 &
  48.5 &
  - &
  0.406 &
  33.9 &
  43.7 &
  53.2 \\ \hline
Emb. &
  RotatE &
  177.1 &
  0.336 &
  24.0 &
  37.2 &
  53.2 &
  258.4 &
  0.401 &
  30.7 &
  45.7 &
  57.1 \\ \hline
\multirow{3}{*}{Rule} &
  NeuralLP &
  1307.3 &
  0.251 &
  \textbf{37.4} &
  27.3 &
  18.8 &
  2204.0 &
  0.255 &
  21.8 &
  26.9 &
  32.4 \\
 &
  DRUM &
  1305.5 &
  0.253 &
  18.9 &
  27.5 &
  37.6 &
  1883.3 &
  0.279 &
  23.6 &
  29.9 &
  35.9 \\
 &
  RNNLogic &
  736.0 &
  0.392 &
  30.5 &
  43.4 &
  56.0 &
  982.2 &
  0.393 &
  33.0 &
  42.2 &
  52.0 \\ \hline
GNN &
  RGCN &
  - &
  0.204 &
  12.4 &
  21.5 &
  36.8 &
  - &
  0.218 &
  13.8 &
  23.9 &
  37.7 \\ \hline
\multirow{2}{*}{GPT-3.5} &
  Inference &
  - &
  - &
  0.4 &
  0.9 &
  2.0 &
  - &
  - &
  23.3 &
  36.6 &
  50.4 \\
 &
  LeSR &
  \textbf{124.3} &
  \textbf{0.420} &
  \underline{32.7} &
  \underline{46.1} &
  59.8 &
  \textbf{95.0} &
  \textbf{0.570} &
  \textbf{45.3} &
  \textbf{64.9} &
  \textbf{77.1} \\ \hline
\multirow{2}{*}{GPT-4} &
  Inference &
  - &
  - &
  3.2 &
  5.0 &
  6.7 &
  - &
  - &
  36.1 &
  48.9 &
  64.8 \\
 &
  LeSR &
  \underline{133.1} &
  0.414 &
  32.0 &
  45.7 &
  \underline{60.2} &
  99.6 &
  \underline{0.549} &
  \underline{44.3} &
  \underline{61.8} &
  73.1 \\ \hline
\multirow{2}{*}{Gemini-1.5} &
  Inference &
  - &
  - &
  0.5 &
  1.0 &
  1.9 &
  - &
  - &
  25.3 &
  39.7 &
  55.8 \\
 &
  LeSR &
  141.4 &
  \textbf{0.420} &
  32.3 &
  \textbf{46.4} &
  \textbf{61.1} &
  \underline{98.5} &
  0.539 &
  43.0 &
  60.7 &
  \underline{73.5} \\ \hline
\end{tabular}
\caption{KBC performance on FB15K-237 and WD15K.  
\label{table: results_34}
}
\end{table*}

To inspect the significance and quality of the learned rule, we borrow the annotation from \cite{lv-etal-2021-multi} which provides rule interpretability scores for WD15K, annotating the most significant reasoning paths based on interpretability and reliability.
More specifically, for each high-confidence rule, they randomly sample up to ten corresponding real paths and label these paths for interpretability, scoring 0, 0.5 or 1.
We define the Rule Clarity Score (RCS) of a single rule as the average interpretability score of its sampled paths, and the RCS of the model being the average RCS of all rules learned by this model. 
We assess models by examining the total number of learned rules, the number of high-confidence rules (those with interpretability annotations), and their RCS, which reflects the interpretability quality of high-confidence rules. To provide a more comprehensive evaluation, we introduce the Rule Quality Index (RQI), which balances the Rule Clarity Score (RCS) and the High-Confidence Rule Ratio (HCR). Inspired by the F1-score metric, RQI measures a model’s ability to generate a high proportion of high-confidence rules while maintaining strong interpretability: $$RQI = 2\cdot \frac{HCR \cdot RCS}{HCR+RCS}\times 100\%,$$ where $$HCR=\frac{\# High\ Confidence\ Rules}{\# Learned\ Rules}\times 100\%.$$

\paragraph{Model and Baseline} 
Under our proposed \textbf{LeSR} framework, we consider three different LLMs in the Proposers: GPT-3.5, GPT-4 and Gemini-1.5.

We choose representative baseline models spanning over different paradigms for the task of KBC. 
For embedding-based methods, we choose the popular \textbf{RotatE} \cite{Sun2018RotatEKG} model, which can implicitly learn simple composition rules, symmetric rules and inverse rules;
For traditional non-ML mining methods, we consider \textbf{AnyBURL} which efficiently mines logical rules from knowledge bases using a bottom-up strategy \cite{Meilicke2019AnytimeBR};
For rule-based methods, we consider \textbf{NeuralLP} \cite{Yang2017DifferentiableLO} that uses a differentiable process for training logical rules with gradient-based learning,  \textbf{DRUM}, which extends NeuralLP by incorporating reinforcement learning to optimize rule discovery \cite{Sadeghian2019DRUMED}, as well as  \textbf{RNNLogic} \cite{Qu2020RNNLogicLL}, a probabilistic approach that trains a rule generator and a reasoning predictor using the EM algorithm and uses RotatE to improve the modelling of reasoning paths;
For GNN-based methods, we incorporate \textbf{RGCN} \cite{Schlichtkrull2017ModelingRD}, which leverages GNNs to model relational data for performing message passing across the graph structure.

We also include a baseline approach which directly leverages an LLM to perform KBC inference, denoted as \textbf{Inference}. Specifically, for each $(h, r, ?)$ query, we extract a subject entity $h$-centered subgraph and use it as contextual information to prompt the LLMs to generate up to 10 most likely object entity candidates. We have tried different prompt templates and have observed non-significant variations in performance, and the final template can be found in Appx.~\ref{appendix: prompt templates}. We use GPT-3.5, GPT-4 and Gemini-1.5 as the LLM backbones.

\paragraph{Implementation}

For the Subgraph Extractor, we use the following configuration. To control the size of the subgraph and LLM context length, we set the maximum steps of multi-hop traversal to three and set the maximum number of neighbouring entities to 3. We sample at most 30 subgraphs for each relation. For the LLM Proposer, we utilize the APIs of corresponding LLMs, namely GPT-3.5, GPT-4 and Gemini-1.5. We put the subgraph and the target triplet into a curated prompt template and fed them into the LLMs to generate the rules. The detailed prompt template for LLMs to propose logic rules can be found in Appx.~\ref{appendix: prompt templates}.

When learning the score for each rule, all model variants use the AdamW optimizer and share an initial learning rate of 0.001 and a weight decay factor of 0.1. We use a StepLR scheduler with a step size of 100 and a gamma value of 100. An early stopping criteria of 30 epochs is further adopted. For experiments on WD15K, we train all models using the same configurations as FB15K-237 following \citet{lv-etal-2021-multi}. 
The implementation details of baseline models can be found in Appx.~\ref{appendix: baseline implementation}

\begin{table*}[htp]
\centering
\small 
\setlength{\tabcolsep}{1.5mm}
\begin{tabular}{c|ccccc|ccccc}
\hline
\multirow{2}{*}{Variation} &
  \multicolumn{5}{c|}{UMLs} &
  \multicolumn{5}{c}{WN18RR} \\
 &
  MR $\downarrow$ &
  MRR$\uparrow$ &
  H@1$\uparrow$ &
  H@3$\uparrow$ &
  H@10$\uparrow$ &
  MR $\downarrow$ &
  MRR$\uparrow$ &
  H@1$\uparrow$ &
  H@3$\uparrow$ &
  H@10$\uparrow$ \\ \hline
GPT-3.5 w/ WL &
  \textbf{5.1} &
  \textbf{0.722} &
  \textbf{64.6} &
  \textbf{75.7} &
  \textbf{86.0} &
  1989.0 &
  \textbf{0.497} &
  \textbf{44.1} &
  \textbf{52.3} &
  \textbf{60.9} \\
GPT3.5 w/o WL &
  10.6 &
  0.364 &
  22.4 &
  40.6 &
  67.7 &
  1985.9 &
  0.460 &
  38.0 &
  50.7 &
  60.5 \\ \hline
GPT-4 w/ WL &
  \underline{5.8} &
  \underline{0.674} &
  \underline{58.6} &
  \underline{71.5} &
  \underline{83.6} &
  1987.8 &
  0.387 &
  29.5 &
  42.8 &
  58.1 \\
GPT-4 w/o WL &
  10.7 &
  0.328 &
  19.3 &
  34.8 &
  64.6 &
  \textbf{1949.4} &
  0.406 &
  31.2 &
  45.4 &
  59.6 \\ \hline
Gemini-1.5 w/ WL &
  7.8 &
  0.640 &
  57.3 &
  65.9 &
  75.3 &
  1987.8 &
  \underline{0.489} &
  \underline{43.0} &
  \underline{51.7} &
  \underline{60.9} \\
Gemini-1.5 w/o WL &
  11.5 &
  0.321 &
  18.1 &
  34.6 &
  63.2 &
  \underline{1951.5} &
  0.476 &
  40.8 &
  51.1 &
  60.7 \\ \hline
\end{tabular}
\caption{Performance on UMLs and WN18RR with and without weight learning. 
\label{table: ablation_uniform_weight_12}
}
\end{table*}

\begin{table*}[htp]
\centering
\small
\setlength{\tabcolsep}{1.5mm}

\begin{tabular}{c|ccccc|ccccc}
\hline
\multirow{2}{*}{Variation} & \multicolumn{5}{c|}{FB15K-237}                       & \multicolumn{5}{c}{WD15K}                          \\
 &
  MR $\downarrow$ &
  MRR$\uparrow$ &
  H@1$\uparrow$ &
  H@3$\uparrow$ &
  H@10$\uparrow$ &
  MR $\downarrow$ &
  MRR$\uparrow$ &
  H@1$\uparrow$ &
  H@3$\uparrow$ &
  H@10$\uparrow$ \\ \hline
GPT-3.5 w/ WL &
  \textbf{122.0} &
  \textbf{0.412} &
  \textbf{32.3} &
  \textbf{45.1} &
  \textbf{58.0} &
  \textbf{97.0} &
  \textbf{0.553} &
  \textbf{44.6} &
  \textbf{62.2} &
  \textbf{74.0} \\
GPT3.5 w/o WL              & \underline{128.9} & 0.374 & 28.1 & 41.7 & 55.1          & 98.4          & 0.520 & 41.7          & 57.4 & 71.6 \\ \hline
GPT-4 w/ WL &
  134.8 &
  \underline{0.401} &
  \underline{31.3} &
  \underline{43.7} &
  56.9 &
  99.8 &
  \underline{0.537} &
  \underline{43.3} &
  \underline{60.2} &
  \underline{72.1} \\
GPT-4 w/o WL               & 136.4          & 0.355 & 26.7 & 38.5 & 52.7          & 98.6          & 0.523 & 42.1          & 57.7 & 71.6 \\ \hline
Gemini-1.5 w/ WL           & 144.1          & 0.397 & 30.9 & 43.4 & \underline{56.9} & 102.8         & 0.522 & 42.2 & 57.9 & 70.1 \\
Gemini-1.5 w/o WL          & 140.0          & 0.354 & 26.5 & 38.9 & 52.6          & \underline{97.4} & 0.511 & 40.5          & 56.9 & 71.6 \\ \hline
\end{tabular}
\caption{Performance on FB15K and WD15K with and without weight learning. 
\label{table: ablation_uniform_weight_34}
}
\end{table*}

\subsection{Result and Analysis}

\begin{table}[htp]
\centering
\scriptsize
\setlength{\tabcolsep}{1.0mm}
\begin{tabular}{c|c|ccccc}
\hline
\multirow{2}{*}{Category}   & \multirow{2}{*}{Model} & \multicolumn{5}{c}{CN100}                                                \\
                            &                        & MR $\downarrow$ & MRR$\uparrow$  & H@1$\uparrow$ & H@3$\uparrow$ & H@10$\uparrow$ \\ \hline
Non-ML                      & AnyBURL                & -               & 0.221          & 14.3          & 24.4          & 38.2           \\ \hline
Emb.                        & RotatE                 & \textbf{4033.7}          & 0.317          & 19.4          & 38.0          & 55.5           \\ \hline
\multirow{3}{*}{Rule}       & NeuralLP               & -               & -              & -             & -             & -              \\
                            & DRUM                   & -               & -              & -             & -             & -              \\
                            & RNNLogic               & 11583.9         & 0.076          & 3.3           & 8.3           & 16.2           \\ \hline
GNN                         & RGCN                   & -               & 0.101          & 4.3           & 10.5          & 21.9           \\ \hline
\multirow{2}{*}{GPT-3.5}    & Inference              & -               & -              & 10.8          & 17.5          & 27.1           \\
                            & LeSR                   & 5866.5          & \textbf{0.345} & \textbf{22.7} & \textbf{41.9} & \textbf{56.7}  \\ \hline
\multirow{2}{*}{GPT-4}      & Inference              & -               & -              & 10.0          & 15.7          & 29.6           \\
                            & LeSR                   & 5873.1          & 0.336          & 22.7          & 39.5          & 54.3           \\ \hline
\multirow{2}{*}{Gemini-1.5} & Inference              & -               & -              & 16.7          & 25.2          & 39.7           \\
                            & LeSR                   & \underline{5840.5}          & \underline{0.344} & \underline{23.0} & \underline{41.2} & \underline{55.3}  \\ \hline
\end{tabular}
\caption{KBC performance on CN100. 
\label{table: results_5}
}
\end{table}

\paragraph{Comparision with Existing KBC Methods} We present the KBC performance results for smaller more niche datasets UMLs and WN18RR in Tab.~\ref{table: results_12}, for larger more realistic datasets FB15K-237 and WD15K in Tab.~\ref{table: results_34} and results for commonsense dataset CN100 in Tab.~\ref{table: results_5}.

We can observe that LeSR achieves highly competitive results across all five datasets from diverse knowledge domains. On smaller, relatively niche datasets UMLs and WN18RR, our models consistently rank among the top three across all metrics. The differences between our approach and pure rule-learning models (NeuralLP and DRUM) are marginal, while the top three methods maintain a substantial lead over the remaining baselines. On larger, more realistic datasets like FB15K-237, WD15K, and CN100, LeSR outperforms all selected baseline models across all evaluation metrics

Compared to the selected pure embedding (RotatE) and pure logic rule (NeuralLP) models, LeSR excels by effectively combining the strengths of both approaches.
Against RNNLogic, which integrates logical rules with knowledge embeddings, our models show comparable performance across all datasets without excessive training, indicating their robustness and reliability when combining the complementary strength of embeddings and logical rules for KBC.
Additionally, our models surpass RGCN across all five datasets, indicating that they leverage structural information more effectively.
When compared to Inference, the superior performance of LeSR highlights the necessity of learning and refining logical rules rather than relying on relevant subgraphs as LLM context.

We notice that certain baselines show considerable performance drops on the CN100K dataset when compared with other datasets. We speculate this is due to CN100K being a commonsense KB where not all facts are explicitly known and, therefore, violates the closed-world assumption these baselines rely on. In fact, NeuralLP and DRUM both require that all entities in the test set are also in the train set, which CN100K does not satisfy.

\paragraph{Impact of LLM Backbone}

\begin{figure}[htp]
\centering
\includegraphics[width=0.95\columnwidth,trim=0cm 0.9cm 0cm 0cm]{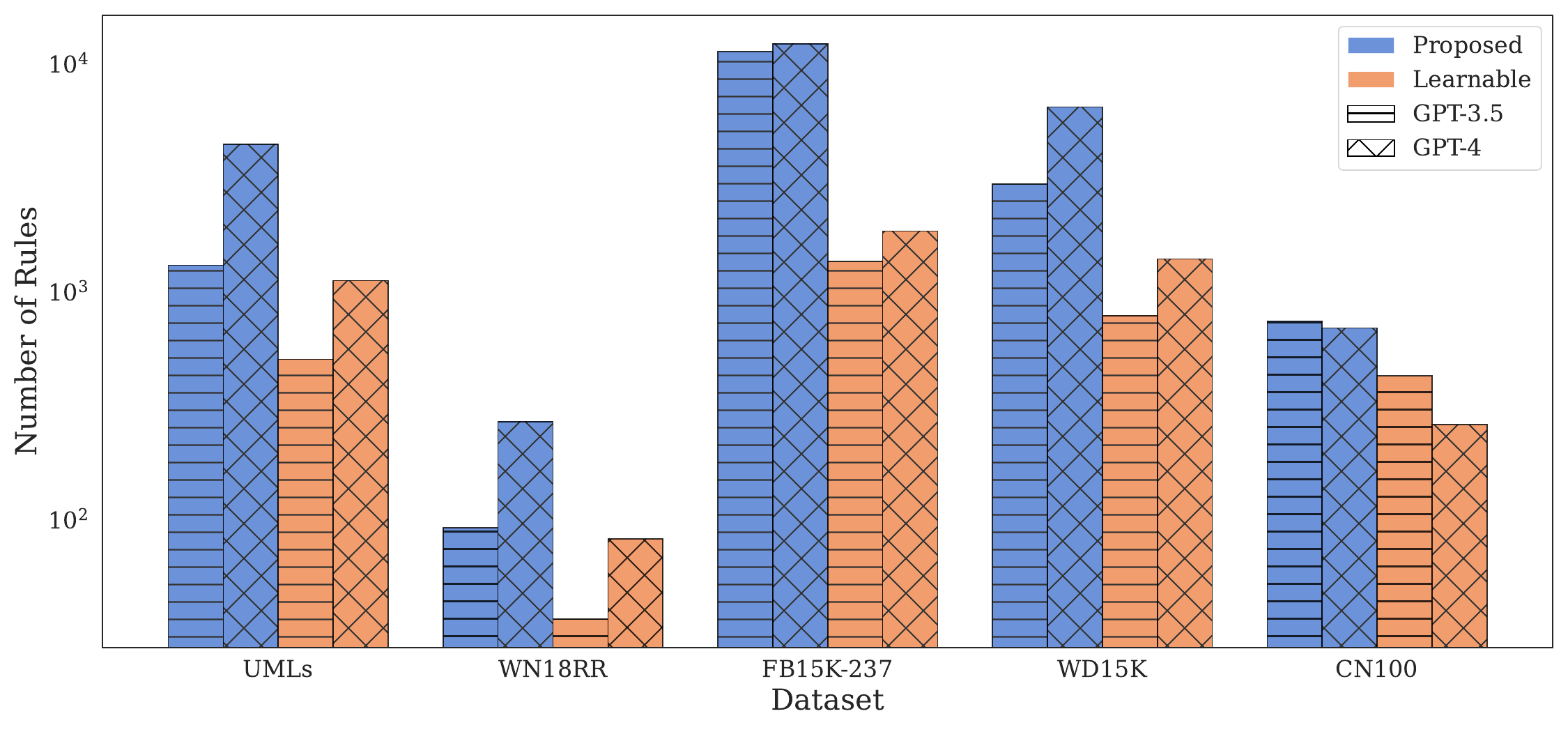} 
\caption{Numbers of proposed and learnable rules using GPT-3.5 and GPT-4. The y-axis is in log 10.}
\label{figure: gpt models proposed vs learnable rule count}
\end{figure}

Surprisingly, switching from GPT-3.5 to GPT-4 does not yield a performance boost except on the UMLs dataset. Upon closer inspection, we notice that using GPT-4 as the LLM backbone in the Proposer leads to significantly more logical rule candidates than using GPT-3.5, as shown in Fig.~\ref{figure: gpt models proposed vs learnable rule count}. However, the additional rules proposed by GPT-4 tend to be patterns restricted to the knowledge subgraph included in the LLM prompt. This means the models have to additionally learn the significance of ungeneralizable rules, thus impacting the KBC performance.

\paragraph{Learning of Rule Signifcance}

To investigate the impact of rule significance learning, we conduct experiments so that the model is trained without weight learning (``WL''), i.e., assigns equal values to $w_i$.

\begin{table}[htp]
\centering
\small
\setlength{\tabcolsep}{0.5mm}
\begin{tabular}{c|ccccc}
\hline
\multirow{2}{*}{Variation} & \multicolumn{5}{c}{CN100}                                                 \\
                           & MR $\downarrow$ & MRR$\uparrow$  & H@1$\uparrow$ & H@3$\uparrow$ & H@10$\uparrow$ \\ \hline
GPT-3.5 w/ WL              & 5867.2          & \textbf{0.342} & \textbf{23.2} & \textbf{40.1} & \textbf{55.5}  \\
GPT3.5 w/o WL              & 5879.6          & 0.190          & 9.8           & 20.5          & 40.7           \\ \hline
GPT-4 w/ WL                & 5870.9          & 0.328          & \underline{23.0} & 36.7          & 52.8           \\
GPT-4 w/o WL               & 5877.7          & 0.226          & 13.0          & 25.1          & 44.0           \\ \hline
Gemini-1.5 w/ WL           & \textbf{5835.3} & \underline{0.339} & 22.7          & \textbf{40.1} & \underline{54.2}  \\
Gemini-1.5 w/o WL          & \underline{5861.2}          & 0.192          & 9.3           & 22.3          & 41.0           \\ \hline
\end{tabular}
\caption{Performance on CN100 with and without weight learning, 
\label{table: ablation_uniform_weight_5}
}
\end{table}

As shown in Tab.~\ref{table: ablation_uniform_weight_12}, ~\ref{table: ablation_uniform_weight_34} and~\ref{table: ablation_uniform_weight_5}, we can see that for models without learning rule significance, the inference results are worse than models with reasoner for all five KB datasets, although the performance gap is relatively small on WN18RR and WD15K. We speculate these may be due to (1) the relatively low quality of logical rules proposed for the WN18RR dataset so that the embedding contributes more to the inference results, and (2) the relatively high quality of LLMs proposed for WD15K, so assigning uniform weights will not introduce too many false deduction results when performing KBC.

\paragraph{Rule Interpretability}

Apart from LeSR, NeuralLP and RNNLogic can explicitly learn logical rules that can be used to perform KBC. We evaluate the quality of learned logical rules using the rule interpretability annotations from WD15K.

\begin{figure}[htp]
\centering
\includegraphics[width=0.95\columnwidth,trim=0cm 0.9cm 0cm 0cm]{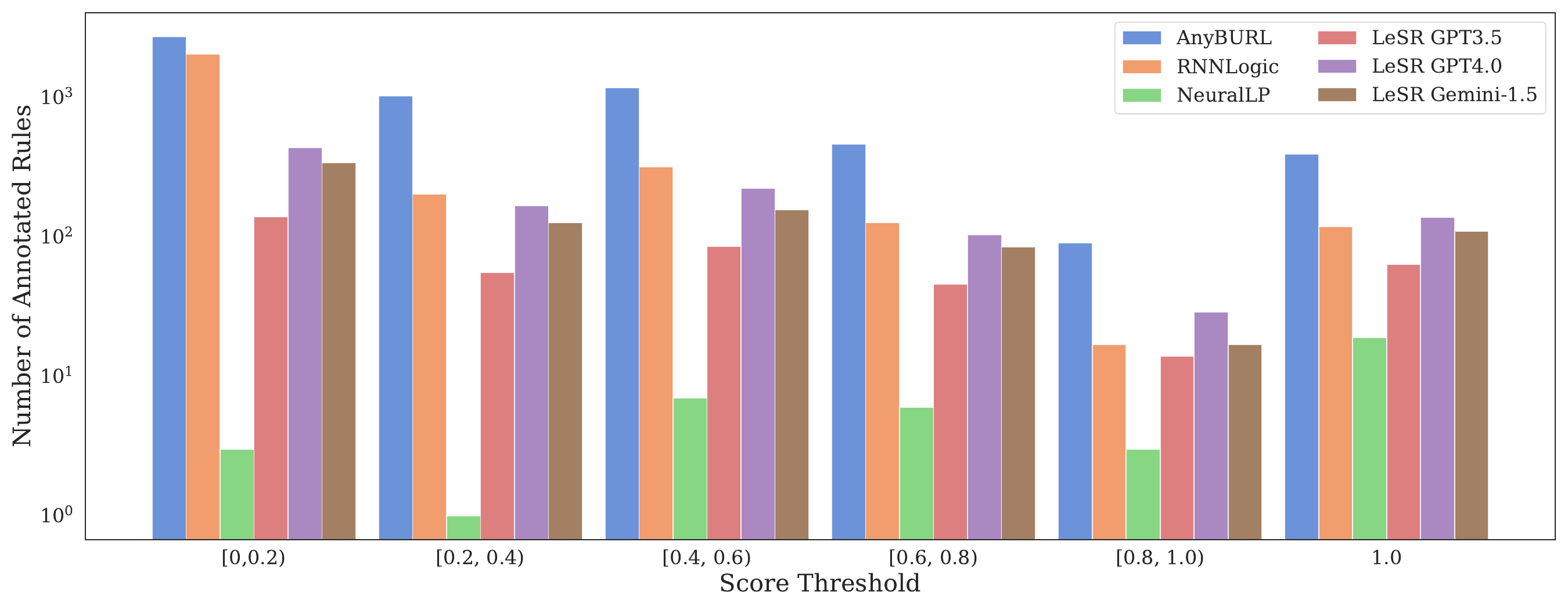} 
\caption{Numbers of learned logical rules of different interpretability scores. The y-axis is in log 10.}
\label{figure: wd15k rule interpretability distribution}
\end{figure}

The human annotations provided by \citet{lv-etal-2021-multi} score only the high-confidence rules in WD15K between 0 and 1. As shown in Tab.~\ref{table: rule quality eval}, We evaluate models based on the total number of learned rules, the High-Confidence Rule Ratio (HCR), the Rule Clarity Score (RCS) and the Rule Quality Index (RQI), as defined in Sec.~\ref{paragraph: metrics}.
We also show the distribution of learned rules by their interpretability scores in Fig.~\ref{figure: wd15k rule interpretability distribution}.

\begin{table}[htp]
\centering
\scriptsize  
\begin{tabular}{c|c|c|c|c|c}
\hline
Model & \# Lrnd & \# HiConf & HCR$\uparrow$ & RCS$\uparrow$  & RQI$\uparrow$ \\
\hline
AnyBURL         & 116,938 & 5,930 & 5.07 & 0.305 & 8.70 \\
NeuralLP         & 1,544 & 39 & 2.53& \textbf{0.764} & 4.89 \\
RNNLogic         & 17,900 & 2,489 & 13.91 & 0.219 & 17.02 \\
LeSR GPT-3.5      & 794 & 406 & 51.13 & \underline{0.428} & \underline{46.60} \\
LeSR GPT-4      & 1,406 & 1,106 & \textbf{78.66} & 0.378 & \textbf{51.03} \\
LeSR Gemini-1.5  & 1,657 & 840 & \underline{50.69} & 0.376 & 43.18\\
\hline
\end{tabular}
\caption{Rule quality evaluation on WD15K for AnyBURL, RNNLogic, NeuralLP and LeSR. Here ``HiConf'' means high-confidence rules: logical rules considered to be highly reliable and with interpretability annotation in the dataset.
}
\label{table: rule quality eval}
\end{table}

Among the models, AnyBURL generates the largest number of learnable rules (116,938), but with a low High-Confidence Rule Ratio (HCR) of 5.07\% and a modest Rule Clarity Score (RCS) of 0.305. This indicates that while AnyBURL is highly productive, the quality and interpretability of its rules are relatively limited. Similarly, RNNLogic produces a considerable number of rules (17,900) but achieves a low RCS of 0.219, with only a small proportion of rules being high-confidence. In contrast, NeuralLP, with only 1,544 rules, excels in interpretability with an RCS of 0.764, focusing on producing a compact set of highly interpretable rules.

Our LeSR models, particularly LeSR GPT-4 and LeSR GPT-3.5, show a more balanced performance, both in terms of the number of rules learned and their interpretability. LeSR GPT-4 stands out with the highest HCR of 78.66\% and a solid RCS of 0.378, achieving the highest Rule Quality Index (RQI) of 51.03, demonstrating its ability to generate high-quality, interpretable rules. LeSR GPT-3.5 also performs well with a high HCR of 51.13\% and an RCS of 0.428, resulting in a strong RQI of 46.60.

Overall, the LeSR models strike a commendable balance between the quantity and quality of rules, achieving higher RQI scores than other models, which tend to focus too much on generating large rule sets (e.g., AnyBURL) or on producing a smaller set of highly interpretable rules (e.g., NeuralLP). The LeSR models, especially GPT-4, effectively blance the number of rules learned and the quality of rules learned.

\section{Conclusion}
In this work, we present a novel framework that improves the transparency and reliability of Knowledge Base Completion by combining LLMs with symbolic reasoning. Our approach leverages the linguistic capability and rule-generating ability of LLMs in conjunction with the verifiable reasoning of rule-based approaches.
Experimental results show that our proposed method not only improves accuracy but also provides generalizability across diverse KB datasets.
By bridging the gap between the interpretability of rule-based reasoning and the adaptability of LLMs, our work offers a potential future direction on enhancing the reliability of KBC while providing a transparent and scalable solution for handling large-scale knowledge bases.


\bibliography{custom_acl}

\appendix

\section{Baseline Implementations} \label{appendix: baseline implementation}

For the baseline models, we utilize the official implementations provided by the authors. Specifically, for RotatE\footnote{\url{https://github.com/DeepGraphLearning/KnowledgeGraphEmbedding}},  NeuralLP\footnote{\url{https://github.com/fanyangxyz/Neural-LP}}, DRUM\footnote{\url{https://github.com/alisadeghian/DRUM}} and AnyBURL\footnote{\url{https://web.informatik.uni-mannheim.de/AnyBURL}}, we employed the respective official codebases and ran the models using the best hyperparameters reported by the authors. For the RNNLogic setup, we combined the default hyperparameters provided by RNNLogic\footnote{\url{https://github.com/DeepGraphLearning/RNNLogic}} with the best RotatE configuration from the RotatE models. Lastly, for RGCN, we re-implemented the model in PyTorch based on the original TensorFlow implementation\footnote{\url{https://github.com/tkipf/relational-gcn}} and executed the models using the best hyperparameters reported by the authors.

\section{Example Rules Learned} \label{appendix: learned rules}
Here we present some logical rules learned by the Rule Reasoner, followed by their rule significance score in percentage.

\paragraph{Relation: ``separated from'' ($\alpha=0.980$)}
\begin{itemize}[noitemsep,topsep=0pt]
\item ``IF (A, shares border with, B) AND (B, separated from, C) THEN (A, separated from, C)'' ($0.32\%$)
\item ``IF (A, country, B) AND (B, shares border with, C) THEN (A, separated from, C)'' ($0.43\%$)
\item ``IF (A, country, B) AND (B, diplomatic relation, C) AND (C, shares border with, D) THEN (A, separated from, D)'' ($25.66\%$)
\item ``IF (A, country, B) AND (B, shares border with, C) AND (C, diplomatic relation, D) THEN (A, separated from, D)'' ($0.25\%$)
\item ``IF (A, diplomatic relation, B) AND (B, country, C) THEN (A, separated from, C)'' ($0.43\%$)
\item ``IF (A, shares border with, B) AND (B, shares border with, C) THEN (A, separated from, C)'' ($72.48\%$)
\item ``IF (A, shares border with, B) AND (B, country, C) THEN (A, separated from, C)'' ($0.43\%$)
\end{itemize}

\paragraph{Relation: ``performer'' ($\alpha=0.878$)}
\begin{itemize}[noitemsep,topsep=0pt]
\item ``IF (X, creator, Y) AND (Y, professional or sports partner, Z) THEN (X, performer, Z)'' ($0.21\%$)
\item ``IF (A, present in work, B) AND (B, cast member, C) THEN (A, performer, C)'' ($99.80\%$)
\end{itemize}

\paragraph{Relation: ``language used'' ($\alpha=0.707$)}
\begin{itemize}[noitemsep,topsep=0pt]
\item ``IF (A, diplomatic relation, B) AND (B, official language, C) THEN (A, language used, C)'' ($99.29\%$)
\end{itemize}

\section{LLM Prompt Templates} \label{appendix: prompt templates}

The bolded parts are places to insert relevant subgraphs and target triplets.

\paragraph{Prompt Template for Rule Proposing} 

The following template is designed to guide LLMs to generate logical rules that can deduce a specific target fact $(h,r,t)$ based on a given knowledge subgraph. The prompt provides example rules to illustrate the format where logical operators are used to describe how multiple conditions involving entities lead to the deduction of a new triplet.

\begin{center}
\fbox{
\parbox{0.9\linewidth}{
A knowledge subgraph describes relationships between entities using a set of triplets. Each triplet is written in the form of triplet (SUBJ, REL, OBJ), which states that entity SUBJ is of relation REL to entity OBJ.

A logic rule can be applied to known triplets to deduce new ones. Each rule is written in the form of a logical implication, which states that if the conditions on the right-hand side are satisfied, then the statement on the left-hand side holds true. Here are some example rules where A, B, C are entities:

IF (A, parent, B) AND  NOT (A, father, B) THEN (A, mother, B)

IF (A, father, B) OR (A, mother, B) THEN (A, parent, B)

IF (A, mother, B) AND (A, sibling, C) THEN (C, mother, B)

Now we have the following triplets:
\textbf{\textit{knowledge subgraph}}

Please generate as many of the most important logical rules based on the above knowledge subgraph to deduce triplet \textbf{\textit{target fact}}. The rules provide general logic implications instead of using specific entities. Return the rules only without any explanations.
}
}
\end{center}

\paragraph{Prompt Template for Direct KBC Inferring}

This prompt template is designed to instruct LLMs to predict the most likely entity candidates that can complete a given query $(h,r,?)$ query given a list of triplets from its relevant knowledge subgraph. We use the generated entity candidates to compute Hit@K metrics for evaluation.

\begin{center}
\fbox{\parbox{0.9\linewidth}{
A knowledge subgraph describes relationships between entities using a set of triplets. Each triplet is written in the form of triplet (SUBJ, REL, OBJ), which states that entity SUBJ is of relation REL to entity OBJ.

Now we have the following triplets:

\textbf{\textit{relevant knowledge subgraph}}

Please generate 10 most likely OBJ candidates to complete \textbf{\textit{query}}. Return only the entity candidates without any additional text.
}}
\end{center}
\section{Logic Rules by Traversal Structure} \label{appendix: rule traversal structures}

The logical rule structures are categorized into three main groups based on the complexity and traversal structure of relations: two-node (0th-order), three-node (1st-order), and four-node (2nd-order) cases. These categories encapsulate different relational patterns, where relation direction and node connectivity play crucial roles in defining the logic. 
Each logical rule structure can be represented using a matrix-based approach, where $M_r$ denotes the matrix for $(h,t)$ pairs satisfying relation $r$. The computation of $\mathbf{C}$, representing the satisfiability of the rule body, involves matrix multiplication for conjunction ($M_{r_i}M_{r_j}$ represents $(h,t)$ pairs so that there exists some $e\in\mathcal{E}$ so that $(h,r_i,e),(e,r_j,t)\in\mathcal{G}$) and matrix transpose for relation inversion ($M_r^\top$ represents $(h,t)$ pairs so that $(t,r,h)\in\mathcal{G}$). Elementwise multiplication $\circ$ is used to compute $\mathbf{A}$ by evaluating $(h,t)$ pairs against the rule head.

\paragraph{0th-order Structures} These two cases represent simple rules involving two entities with direct or inversed relations.

\textbf{Case 0-1}: $(A, r_i, B) - (A, r_j, B)$ with example rule $\phi=$``IF $(A,\ place\ of\ birth,\ B))$ THEN $(A,\ country\ of\ citizenship,\ B)$'':
$$\mathbf{C} = M_{r_i} \text{, and }   \mathbf{A} = \mathbf{C} \circ M_{r_j}$$

\textbf{Case 0-2}: $(A, r_i, B) - (B, r_j, A)$ with example rule $\phi=$``IF $(A,\ award\ received,\ B)$ THEN $(B,\ winner,\ A)$'':
$$\mathbf{C} = M_{r_i}^\top\text{, and }  \mathbf{A} = \mathbf{C} \circ M_{r_j}$$

\paragraph{1st-order Structures} The following cases introduce a third entity node, adding complexity with various chain structures that either converge or diverge, providing greater diversity in logical structures.

\textbf{Case 1-1}: $(A, r_i, B) -  (B, r_j, C) - (A, r_k, C)$ with example rule $\phi=$``IF $(A,\ genre,\ B)$ AND $(B,\ subclass\ of,\ C)$ THEN $(A,\ genre,\ C)$'':
$$\mathbf{C} = M_{r_i}M_{r_j} \text{, and }   \mathbf{A} =\mathbf{C} \circ M_{r_k}$$

\textbf{Case 1-2}: $(B, r_i, A) - (B, r_j, C) - (A, r_k, C)$ with example rule $\phi=$``IF $(B,\ composer,\ A)$ AND $(B,\ country\ of\ origin,\ C)$ THEN $(A,\ country\ of\ citizenship,\ C)$'':
$$\mathbf{C} = M_{r_i}^\top M_{r_j} \text{, and }   \mathbf{A} =\mathbf{C} \circ M_{r_k}$$

\textbf{Case 1-3}: $(A, r_i, B) - (C, r_j, B) - (A, r_k, C)$ with example rule $\phi=$``IF $(A,\ nominated\ for,\ B)$ AND $(C,\ award\ received,\ B)$ THEN $(A,\ performer,\ C)$'':
$$\mathbf{C} = M_{r_i}M_{r_j}^\top \text{, and }   \mathbf{A} =\mathbf{C} \circ M_{r_k}$$

\textbf{Case 1-4}: $(B, r_i, A) - (C, r_j, B) - (A, r_k, C)$ with example rule $\phi=$``IF $(A,\ owner\ of,\ B)$ AND $(B,\ parent\ organization,\ C)$ THEN $(C,\ owned\ by,\ A)$'':
$$\mathbf{C} = M_{r_i}^\top M_{r_j}^\top \text{, and }   \mathbf{A} =\mathbf{C} \circ M_{r_k}$$

\paragraph{2nd-order Structures} These four-node structures exhibit the highest structural diversity. Eight distinct patterns include sequences of relations that form complex structures that cannot be easily captured using simple chain and relation inversion.

\textbf{Case 2-1}: $(A, r_i, B) -  (B, r_j, C) - (C, r_k, D) - (A, r_l, D)$ with example rule $\phi=$``IF $(A,\ twinned\ administrative\ body,\ B)$ AND $(B,\ twinned\ administrative\ body,\ C)$ AND $(C,\ country,\ D)$ THEN $(A,\ country,\ D)$'':
$$\mathbf{C} = M_{r_i} M_{r_j} M_{r_k} \text{, and }   \mathbf{A} =\mathbf{C} \circ M_{r_l}$$

\textbf{Case 2-2}: $(A, r_i, B) -  (B, r_j, C) - (D, r_k, C) - (A, r_l, D)$ with example rule $\phi=$``IF $(A,\ diplomatic\ relation,\ B)$ AND $(B,\ continent,\ C)$  AND  $(D,\ continent,\ C)$ THEN $(A,\ shares\ border\ with,\ D)$'':
$$\mathbf{C} = M_{r_i} M_{r_j} M_{r_k}^\top  \text{, and }   \mathbf{A} =\mathbf{C} \circ M_{r_l}$$

\textbf{Case 2-3}: $(A, r_i, B) -  (C, r_j, B) - (C, r_k, D) - (A, r_l, D)$ with example rule 
$\phi=$``IF $(A,\ country\ for\ sport,\ B)$ AND $(C,\ country\ of\ citizenship,\ B)$ AND $(C,\ participant\ of,\ D)$ THEN $(A,\ participant\ of,\ D)$'':

$$\mathbf{C} = M_{r_i} M_{r_j}^\top M_{r_k}  \text{, and }   \mathbf{A} =\mathbf{C} \circ M_{r_l}$$

\textbf{Case 2-4}: $(B, r_i, A) -  (B, r_j, C) - (C, r_k, D) - (A, r_l, D)$ with example rule $\phi=$`` IF $(A,\ country,\ B)$ AND $(A,\ shares\ border\ with,\ C)$ AND $(C,\ located\ on\ terrain\ feature,\ D)$ THEN $(B,\ located\ on\ terrain\ feature,\ D)$'':
$$\mathbf{C} = M_{r_i}^\top M_{r_j} M_{r_k} \text{, and }   \mathbf{A} =\mathbf{C} \circ M_{r_l}$$

\textbf{Case 2-5}: $(A, r_i, B) -  (C, r_j, B) - (D, r_k, C) - (A, r_l, D)$ with example rule $\phi=$`` IF $(A,\ participant,\ B)$ AND $(C,\ diplomatic\ relation,\ B)$ AND $(D,\ diplomatic\ relation,\ C)$ THEN $(A,\ participant,\ D)$'':
$$\mathbf{C} = M_{r_i} M_{r_j}^\top M_{r_k}^\top \text{, and }   \mathbf{A} =\mathbf{C} \circ M_{r_l}$$

\textbf{Case 2-6}: $(B, r_i, A) -  (B, r_j, C) - (D, r_k, C) - (A, r_l, D)$ with example rule $\phi=$ ``IF $(B,\ characters,\ A)$ AND $(B,\ cast\ member,\ C)$ AND $(D,\ performer,\ C)$ THEN $(A,\ partner,\ D)$'':
$$\mathbf{C} = M_{r_i}^\top M_{r_j} M_{r_k}^\top \text{, and }   \mathbf{A} =\mathbf{C} \circ M_{r_l}$$

\textbf{Case 2-7}: $(B, r_i, A) -  (C, r_j, B) - (C, r_k, D) - (A, r_l, D)$ with example rule $\phi=$ ``IF $(B,\ diplomatic\ relation,\ A)$ AND $(C,\ diplomatic\ relation,\ B)$ AND $(C,\ currency,\ D)$ THEN $(A,\ currency,\ D)$'':
$$\mathbf{C} = M_{r_i}^\top M_{r_j}^\top M_{r_k} \text{, and }   \mathbf{A} =\mathbf{C}\circ M_{r_l}$$

\textbf{Case 2-8}: $(B, r_i, A) -  (C, r_j, B) - (D, r_k, C) - (A, r_l, D)$ with example rule $\phi=$ ``IF $(B,\ diplomatic\ relation,\ A)$ AND $(C,\ country,\ B)$ AND $(D,\ educated\ at,\ C)$ THEN $(A,\ head\ of\ government,\ D)$'': 
$$\mathbf{C} = M_{r_i}^\top M_{r_j}^\top M_{r_k}^\top \text{, and } \mathbf{A} =\mathbf{C} \circ M_{r_l}$$
\end{document}